\documentclass[letterpaper]{article} 
\usepackage{aaai2027}  
\usepackage{amsmath}
\usepackage{amssymb}
\usepackage[hyphens]{url}  
\usepackage{graphicx} 
\usepackage{natbib}  
\usepackage{caption} 
\usepackage{booktabs}
\usepackage{algorithm}
\usepackage{algpseudocode}
\usepackage{multirow}
\usepackage[table]{xcolor} 
\usepackage{arydshln}
\usepackage{placeins}
\usepackage{tabularx}   
\usepackage{xcolor}     
\usepackage{booktabs}   
\usepackage{makecell}   

\newcommand{\best}[1]{\textbf{#1}}
\newcommand{\second}[1]{\underline{#1}}

\title{RailSyn: Diagnosis-Guided Image Generation for Traceable Data Completion in Railway Foreign Object Detection}

\author{
    Quan Hao\textsuperscript{\rm 1},
    Chenxi Zhang\textsuperscript{\rm 1},
    Ziyang Tao\textsuperscript{\rm 2},
    Yuyuan Zhou\textsuperscript{\rm 3,\rm 4},
    Yudong Wang\textsuperscript{\rm 6},
    Rui Shi\textsuperscript{\rm 1},
    Lechuan Xu\textsuperscript{\rm 5},\\
    Liu Changhao\textsuperscript{\rm 1},
    Liguo Zhang\textsuperscript{\rm 1}
}
\affiliations{
    \textsuperscript{\rm 1}The School of Information Science and Technology, Beijing University of Technology\\
    \textsuperscript{\rm 2}Pratt School of Engineering, Duke University\\
    \textsuperscript{\rm 3}Academy of Mathematics and Systems Science, Chinese Academy of Sciences\\
    \textsuperscript{\rm 4}School of Mathematical Sciences, University of Chinese Academy of Sciences\\
    \textsuperscript{\rm 5}Industrial Systems Engineering and Management, National University of Singapore\\
    \textsuperscript{\rm 6}Institute of Automation, Chinese Academy of Sciences\\
    haoquan@emails.bjut.edu.cn,
    zcx20041004@emails.bjut.edu.cn,
    ziyang.tao@duke.edu,
    zhouyuyuan@amss.ac.cn,
    luke.xu@u.nus.edu,
    yudong.wang@ia.ac.cn,
    ruishi@bjut.edu.cn,
    liu20050126@emails.bjut.edu.cn,
    zhangliguo@bjut.edu.cn
}

\begin{document}

\maketitle

\begin{abstract}
Railway foreign object detection (RFOD) is critical to safe railway operation, yet scarce real positive samples incompletely represent task-relevant variations in object scale, intrusion relation, railway scene, illumination, and adverse weather. Existing synthetic augmentation can improve RFOD detection, but its gains lack an explicit account of the task-relevant deficiencies complemented by the generated data. We therefore introduce RailSyn, a diagnosis-guided framework comprising a real-referenced Inspector and a requirement-aligned Generator. The Inspector constructs a variable-radius empirical cover from finite real observations to localize candidate completion regions and profile synthetic pools. The resulting audit identifies railway-context, intrusion-semantic, and visual-consistency requirements; the Generator addresses them through domain adaptation, agent-planned placement and physical contact relations, and plan-consistent conditional refinement. Using the Inspector, we further trace representation-space changes across generation variants; the complete system attains a local-shell occupation of $C_{\mathrm{gap}}$ to 13.64\%, which measures generated coverage of real-derived completion regions. Extensive experiments show AP50--95 gains of up to 4.9 points and consistent improvements across nine mainstream detectors, demonstrating broad cross-architecture utility.

\end{abstract}

\begin{figure*}[!t]
    \centering
        \includegraphics[width=1.0\textwidth]{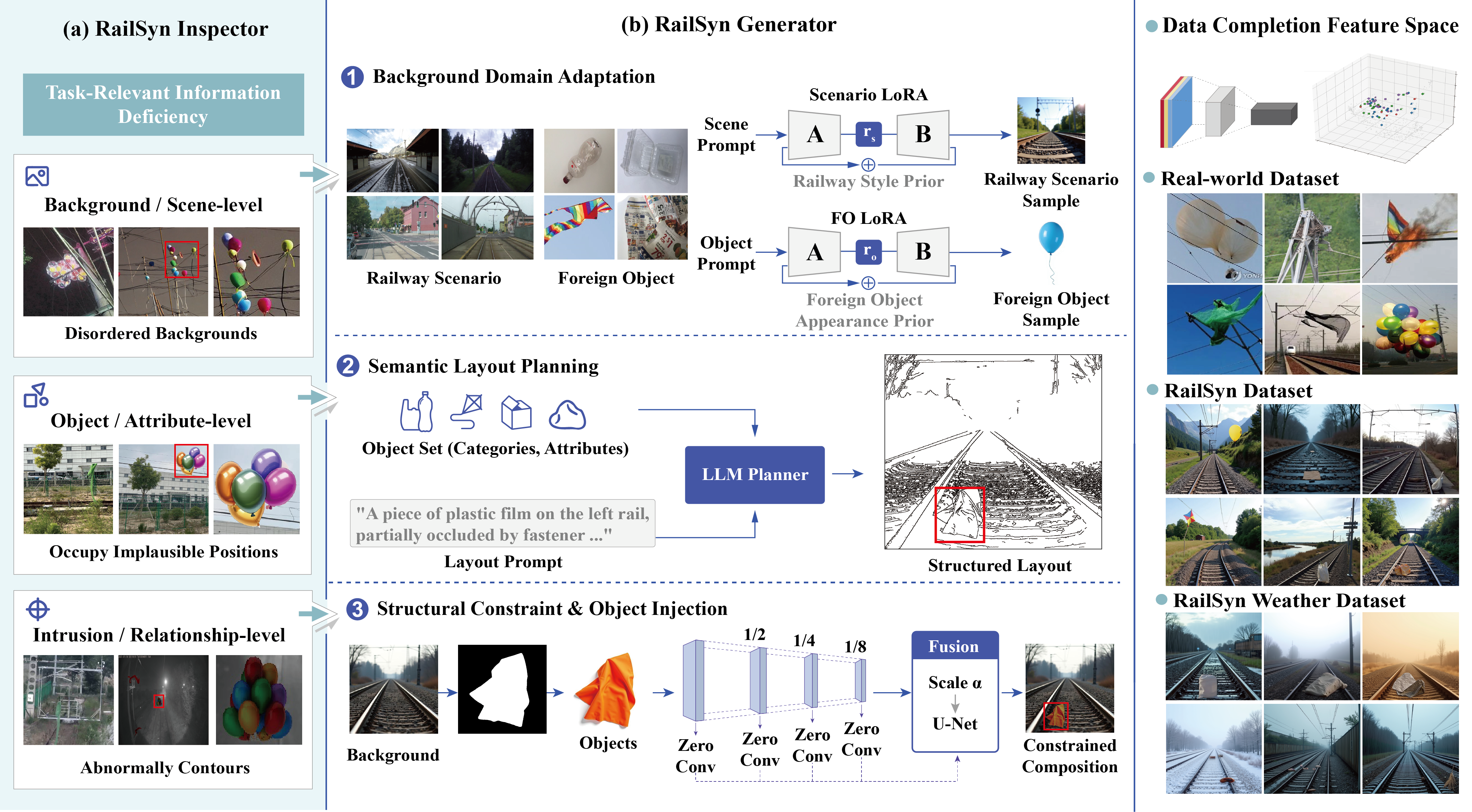}%
\caption{RailSyn overview. (a) Inspector identifies high-uncertainty regions from the finite real reference and inspects nearby SD-based RFOD23 AIGC samples, revealing three recurring patterns. (b) Generator addresses them through railway scene preparation, relation‑aware intrusion planning, and intrusion‑pattern refinement. Right: generated samples and feature‑space completion ($C_{\mathrm{gap}}$).}
\label{fig:fig1}
\end{figure*}

\section{Introduction}
\label{sec:introduction}

Railway foreign object detection (RFOD) is critical to the safe operation of high-speed rail systems \cite{hao2026generative,chen2024railfod23}. Reliable detection requires examples that jointly represent railway structures, environmental conditions, object scales, and the physical relations by which foreign objects intrude into operational space \cite{hao2026generative,chen2024railfod23,chen2025yoloms,oza2024uda}. Such positive samples are scarce and costly to annotate, while small targets further concentrate task evidence in limited image regions \cite{hao2026generative,chen2024railfod23,cheng2023smallobject,shi2025hsfpn}. Consequently, finite RFOD datasets incompletely capture not only visual appearance, but also the context, intrusion semantics, and local object--background effects that determine whether synthetic samples are useful for detection.

Existing augmentation and generation methods produce additional training samples \cite{hao2026generative,feng2024instagen,wang2024detdiffusion}. Conventional operations are confined to observed objects and contexts \cite{ghiasi2021copypaste,trabucco2024effective}, whereas advanced approaches such as SVDDD, RegionDiffusion, and broader grounded generation frameworks prioritize realism, diversity, or controllability \cite{wang2025svddd,li2025regiondiffusion,li2023gligen}. However, none of these objectives explicitly identify which task-relevant regions remain weakly complemented relative to real RFOD observations; consequently, method selection is decoupled from a concrete completion demand. As a result, generated pools may repeat existing patterns, misalign object--scene intrusion semantics, or exhibit anomalous injection effects \cite{somepalli2023diffusionart,wu2024selfcorrecting,song2024imprint}. Their detection gains are likewise difficult to attribute: global similarity metrics are dominated by railway backgrounds, while AP provides a standardized measure of overall detector performance but still falls short of characterizing the contextual, relational, or localized contributions introduced by generated data \cite{jayasumana2024rethinkingfid,ghosh2023geneval,hu2023tifa,lee2023heim,bolya2020tide}. The outstanding gap, therefore, is a unified framework that diagnoses deficient information prior to synthesis, translates that evidence into explicit generation requirements, and evaluates whether the resulting pool effectively complements the specified regions.

We address this gap with RailSyn, an Inspector--Generator framework in which inspection explicitly guides generation. The Inspector builds a variable-radius empirical cover from our collected high-quality real RFOD reference and evaluates synthetic pools using reliable-support coverage ($C_{\mathrm{rel}}$), local-gap completion ($C_{\mathrm{gap}}$), and nonredundant volume efficiency ($\eta_{\mathrm{vol}}$). It applies the real-derived local-shell criterion underlying $C_{\mathrm{gap}}$ to the SD-based\cite{rombach2022ldm} RFOD23 AIGC pool and retrieves representative samples from the high-uncertainty regions induced by the finite real reference, followed by inspection of nearby RFOD23 samples. As shown on the left of Figure~\ref{fig:fig1}, these samples reveal gaps in railway-background fidelity, object--scene intrusion semantics, and anomalous edge effects around injected objects. Guided by this diagnosis, the Generator aligns one process with each gap: railway-domain low-rank adaptation prepares structural backgrounds \cite{hu2022lora}; a multimodal Agent produces executable background-conditioned placement and physical-contact plans; and plan-consistent conditional refinement translates spatial and relation metadata into a unified control condition with mask-aware composition. RailSyn thus turns Inspector-localized deficiencies into explicit generation requirements for complementing the diagnosed regions and improving detection.

RailSyn data improve AP50--95 across all nine tested mainstream detectors, with an average best observed gain of 3.03 points and a maximum gain of 4.9 points. We further evaluate matched Generator ablations with DEIM and YOLO11, and reapply the Inspector to every ablated pool to trace representation-space changes. The complete Generator attains $C_{\mathrm{gap}}=13.64\%$, and the Inspector diagnostics exhibit trends broadly consistent with the AP changes in the ablation study. This descriptive alignment links detector gains with greater occupation of the real-derived completion regions without treating $C_{\mathrm{gap}}$ as an AP predictor. External-pool experiments further distinguish empirical completion from downstream utilization.

Our main contributions are:
\begin{itemize}
    \item We introduce RailSyn, an Inspector--Generator framework that connects real-referenced deficiency analysis, guided synthesis, and representation-space assessment for traceable RFOD data completion.

    \item We develop a diagnosis-guided Generator that aligns railway-domain adaptation, Agent-planned intrusion semantics, and plan-consistent conditional refinement with the context, relation, and boundary gaps exposed by the Inspector.

    \item We demonstrate cross-architecture utility across nine heterogeneous detectors and conduct controlled DEIM/YOLO11 ablations with Inspector profiles to trace component-level effects on detection performance and local-gap completion.
\end{itemize}

\section{Related Work}

\subsection{Railway Foreign-Object Detection and Data Scarcity}

Railway foreign object detection (RFOD) localizes debris and intrusions on or near railway tracks \cite{hao2026generative,chen2024railfod23}, where missed objects can obstruct operation and threaten safety \cite{hao2026generative,chen2024railfod23}. Current detectors mainly follow YOLO-style dense prediction or DETR-style end-to-end set prediction \cite{redmon2016yolo,carion2020detr}: YOLO prioritizes efficient multi-scale detection \cite{chen2025yoloms}, whereas DETR exploits global context through transformer-based matching \cite{carion2020detr}. Despite this progress, RFOD remains a small-sample problem: rare intrusions and costly acquisition limit positive examples, weakening detector learning across scale, illumination, weather, and intrusion relations \cite{hao2026generative,chen2024railfod23,cheng2023smallobject,oza2024uda,kang2019fewshot}. Data-centric augmentation and generation therefore offer a practical route to supplement scarce RFOD training samples \cite{feng2024instagen,wang2024detdiffusion,islam2024diffusemix}. Conventional transformations and copy--paste reuse observed content and provide limited control over scale, placement, and intrusion relations \cite{ghiasi2021copypaste,trabucco2024effective}, motivating more flexible generative augmentation for the small-sample setting.

\subsection{Generative Augmentation and Traceable Completion}

Latent diffusion enables high-resolution synthesis in compressed spaces and supports data augmentation under limited observations \cite{rombach2022ldm,trabucco2024effective,islam2024diffusemix}. More recently, flow matching and FLUX have established a rectified-flow Transformer paradigm for high-quality generation \cite{lipman2023flowmatching,esser2024rectifiedflow}; spatial controls improve placement and structure preservation \cite{li2023gligen,zhang2023controlnet,jia2024ssmg}, while low-rank adaptation enables efficient domain specialization \cite{hu2022lora}. These advances expand the potential of controllable generation for small-sample detection, complementing few-shot learning, rebalancing, domain adaptation, active selection, and focal reweighting \cite{kang2019fewshot,lin2017focalloss,oza2024uda,wan2024datasetrefinement}. However, realism and controllability do not establish whether generated data complement detection-relevant information \cite{ghosh2023geneval,hu2023tifa,lee2023heim,huang2025t2icompbenchpp}, especially for rare RFOD scenes, scales, weather, and intrusion relations \cite{hao2026generative,chen2024railfod23,cheng2023smallobject,oza2024uda}. Existing evaluation offers limited traceability from generation to detection accuracy \cite{jayasumana2024rethinkingfid,lee2023heim,bolya2020tide}: global distribution metrics such as FID and KID can be dominated by railway backgrounds and underweight the small foreign-object regions that carry detection evidence \cite{jayasumana2024rethinkingfid,cheng2023smallobject,ghosh2023geneval}, whereas AP aggregates utility without identifying the contributions of scene coverage, object attributes, intrusion relations, or detector optimization \cite{bolya2020tide,huang2025t2icompbenchpp}. The remaining gap is therefore traceable generation that links observed data deficiencies, generation choices, synthetic-pool completion, and downstream detection gains.

\section{Method}
\label{sec:method}

RailSyn formulates synthetic RFOD augmentation as the traceable completion of a finite real training set. The Inspector constructs real-referenced empirical completion regions, retrieves existing synthetic candidates for analysis, and profiles the generated pools. The Generator then fulfills the resulting railway-context, intrusion-relation, and visual-integration requirements through three dedicated generation modules. We first define completion quality and real-scene detection utility as separate evaluation targets, and subsequently present the Inspector and the requirement-aligned Generator.

\subsection{Problem Formulation}
\label{sec:problem-formulation}

Let $\mathcal D_R=\{x_i\}_{i=1}^{N}$ denote a finite real RFOD training set, $\mathcal D_G=\{g_j\}_{j=1}^{M}$ a generated pool, and $\mathcal D_V$ a disjoint real-scene evaluation set. We study whether $\mathcal D_G$ complements task-relevant information insufficiently represented by $\mathcal D_R$, such that a detector trained on $\mathcal D_R\cup\mathcal D_G$ generalizes better to $\mathcal D_V$ than one trained on $\mathcal D_R$ alone. This objective entails two distinct questions: what information $\mathcal D_G$ contributes relative to the finite real observations, and whether a downstream detector can exploit that contribution. Accordingly, we assess information completion with the Inspector and real-scene utility with detector AP under a fixed training and evaluation protocol.

\subsection{Inspector: Real-Referenced Data Completion Analysis}
\label{sec:inspector}

The Inspector instantiates the completion objective in a task-relevant representation space. We adopt the recent large-scale Qwen3-VL-Embedding-8B~\cite{li2026qwen3vlembeddingqwen3vlrerankerunifiedframework} as the primary encoder because its visual-semantic modeling capacity and 4096-dimensional representation support fine-grained distinctions among railway context, foreign-object appearance, and object--scene intrusion relations. CLIP and DINOv2 serve as established auxiliary encoders for cross-encoder validation. Within each encoder, $\ell_2$-normalized embeddings are analyzed with spherical distance in the complete native space.

To estimate spatially varying support, let $R=\{z_i\}_{i=1}^{N}$ denote the normalized real embeddings. We assign each anchor a radius given by its spherical distance to a fixed-order nearest real neighbor, producing a cover that contracts in densely observed regions and expands across sparse ones. A local intrinsic dimension estimated from real--real neighborhoods then specifies the corresponding volume law while the original embedding coordinates remain intact. This construction captures the nonuniform geometry of finite RFOD observations and provides the real-only reference for subsequent completion analysis.

The anchor radii characterize observed neighborhoods but leave the regions between real samples unresolved. We therefore augment the anchor set with deterministic spherical midpoints of edges connecting neighboring real samples and fit a real-only Matérn-$5/2$ Gaussian process to the log-radius field, $f(\mathbf{z})=\log r(\mathbf{z})$. For each query point $q$, the posterior mean $\mu_f(q)$ estimates the local reach, while the posterior standard deviation $\sigma_f(q)$ quantifies interpolation uncertainty:
\begin{equation}
\begin{aligned}
\mu_f(q)
&= \bar{f}+k_q^\top\bigl(K+\sigma_n^2I\bigr)^{-1}\cdot\bigl(f-\bar{f}),\\
\sigma_f^2(q)
&= K(q,q)-k_q^\top\bigl(K+\sigma_n^2I\bigr)^{-1}\cdot k_q,\\
r^-(q)
&= \exp\!\bigl(\mu_f(q)-\tau\sigma_f(q)\bigr),\\
\widehat{r}(q)
&= \exp\!\bigl(\mu_f(q)\bigr),\\
r^+(q)
&= \exp\!\bigl(\mu_f(q)+\tau\sigma_f(q)\bigr).
\end{aligned}
\label{eq:gp_radius}
\end{equation}
The Gram matrix $K$ is generated by the positive-definite Matérn-$5/2$ kernel:
\begin{equation}
\begin{aligned}
K(\mathbf{z},\mathbf{z}')
&= \sigma_f^2\left(1+\xi+\frac{\xi^2}{3}\right)e^{-\xi},\\
\xi
&= \frac{\sqrt{5}\,d_c(\mathbf{z},\mathbf{z}')}{\ell},\\
d_c(\mathbf{z},\mathbf{z}')
&= \lVert\mathbf{z}-\mathbf{z}'\rVert_2.
\end{aligned}
\label{eq:matern52_sphere}
\end{equation}

The interpolated radii define a lower reliable support and an upper plausible boundary, while the interval between them exposes localized completion opportunities induced by finite observations. We retain these intervals as query-indexed shells so that overlap between neighboring balls does not erase local demand. After freezing the real construction, each generated sample receives a radius from the posterior mean field:
\begin{equation}
\begin{aligned}
\mathcal S^-&=\bigcup_{q\in Q}\mathcal B_{\mathbb S}(q,r^-(q)),\\
\mathcal S^+&=\bigcup_{q\in Q}\mathcal B_{\mathbb S}(q,r^+(q)),\\
\mathcal G_q&=\mathcal B_{\mathbb S}(q,r^+(q))\setminus
\mathcal B_{\mathbb S}(q,r^-(q)),\\
\mathcal S_G&=\bigcup_{g\in G}\mathcal B_{\mathbb S}(g,\widehat r(g)).
\end{aligned}
\label{eq:confidence-covers}
\end{equation}
Together, $\mathcal S^-$, the indexed family $\{\mathcal G_q\}_{q\in Q}$, and $\mathcal S_G$ provide the set relations needed to assess how generated data retain established support, occupy real-derived gaps, and expand without redundancy.

Let $\mu_{\widehat m}$ denote the local mass induced by the real-only intrinsic-dimensional volume law. The completion profile is defined as
\begin{equation}
\begin{aligned}
C_{\mathrm{rel}}&=\frac{\mu_{\widehat m}(\mathcal S_G\cap\mathcal S^-)}{\mu_{\widehat m}(\mathcal S^-)},\\
C_{\mathrm{gap}}&=\frac{\sum_{q\in Q}\mu_{\widehat m}(\mathcal S_G\cap\mathcal G_q)}{\sum_{q\in Q}\mu_{\widehat m}(\mathcal G_q)},\\
\eta_{\mathrm{vol}}&=\frac{\mu_{\widehat m}(\mathcal S_G)}{\sum_{g\in G}\mu_{\widehat m}(\mathcal B_{\mathbb S}(g,\widehat r(g)))}.
\end{aligned}
\label{eq:cover-diagnostics}
\end{equation}
Here $C_{\mathrm{rel}}$ quantifies reliable-support reach, $C_{\mathrm{gap}}$ quantifies the mass-weighted occupation of indexed local shells, and $\eta_{\mathrm{vol}}$ quantifies generated-cover nonredundancy. Since exact integration over high-dimensional ball unions is intractable, we estimate these masses by native-spherical Monte Carlo: samples are drawn from the local tangent volume law, mapped back to the unit sphere, and weighted by inverse cover multiplicity to avoid double-counting overlaps.

We apply the Inspector to the real reference set to identify regions where finite observations induce the largest empirical uncertainty. For each query $q$, the corresponding shell is ranked by the intrinsic-dimensional mass proxy
\begin{equation}
g(q)\propto \bigl(r^+(q)\bigr)^{\widehat m}-
\bigl(r^-(q)\bigr)^{\widehat m}.
\label{eq:gap-audit}
\end{equation}
High-mass shells define the audit regions, around which spherical nearest-neighbor retrieval collects a real reference and SD-based RFOD23 AIGC candidates. This real-only ranking converts abstract uncertainty regions into directly inspectable image groups while keeping the audit independent of downstream generation choices.

We use the Inspector to analyze high-uncertainty regions induced by the finite real reference, followed by inspection of nearby RFOD23 samples. As shown in Figure~\ref{fig:fig1}(a), the first representative group contains disordered backgrounds that bear little correspondence to authentic railway scenes, indicating a pronounced railway-context deficiency. In the second group, foreign objects occupy implausible positions relative to the track infrastructure, revealing a deficiency in intrusion semantics. In the third group, inserted objects exhibit abnormally sharp contours, conflicting with the weak object--background contrast typical of small foreign objects and exposing a deficiency in intrusion appearance. These three observations motivate the Generator's scene preparation, relation-aware planning, and boundary-aware injection, respectively.

\subsection{Generator: Targeted Synthetic Data Generation}

The Generator translates the three diagnosed deficiencies into a coupled synthesis process. Railway scene preparation addresses context deficiency by injecting railway-scene and foreign-object priors; relation-aware intrusion planning addresses semantic deficiency by conditioning placement on scene geometry and physical relations; and diagnosed intrusion-pattern refinement reconciles object contours with local background appearance. These stages preserve a one-to-one correspondence between Inspector diagnosis and targeted synthesis.

\subsubsection{Railway Scene Preparation}

The context deficiency requires both recognizable railway environments and faithful foreign-object appearance. We therefore learn two low-rank adaptations (LoRAs)~\cite{hu2022lora,blackforest2024flux}: a scene adapter captures track layout, viewpoint, surrounding environment, and illumination, while an object adapter captures category-specific shape, texture, and appearance. During synthesis, the two adapters inject complementary scene and object features into the generation process, providing domain-specific visual priors for high-quality railway composition before spatial and relational constraints are imposed.

\subsubsection{Relation-Aware Intrusion Planning}
\label{sec:agent}

The semantic deficiency arises when foreign-object category and position are specified independently of the selected railway scene. Given a prepared background $I_b$ and a candidate object, a multimodal Agent produces a structured intrusion plan
\begin{equation}
\mathcal Y=\mathcal A(I_b,I_o)
=\{c,b_o,r,u,\theta\},
\label{eq:layout}
\end{equation}
where $c$ is the object category, $b_o$ its placement box, $r$ its physical relation to railway infrastructure, $u$ the associated contact region, and $\theta$ an applicable tilt. The plan explicitly couples placement with rails, sleepers, ballast, or catenary and is deterministically normalized to valid image coordinates and scale ranges. Its accepted box initializes the detector annotation, while its relation and contact fields are propagated to conditional realization. Thus, the Agent converts background--object compatibility into an executable interface between scene preparation and image synthesis.

\begin{table*}[ht]
\centering
\caption{Cross-architecture detection performance with increasing numbers of RailSyn samples (AP50/AP50--95, \%).}
\label{tab:mega60}
\footnotesize
\setlength{\tabcolsep}{4pt}
\begin{tabularx}{\textwidth}{
  @{}
  l
  >{\centering\arraybackslash}X
  >{\centering\arraybackslash}X
  >{\centering\arraybackslash}X
  >{\centering\arraybackslash}X
  >{\centering\arraybackslash}X
  >{\centering\arraybackslash}X
  @{}
}
\toprule
Detector & Real & $+$100 & $+$200 & $+$300 & $+$400 & $+$500 \\
\midrule
DEIM      & 72.0/46.2 & 75.4/48.3 & 78.2/48.6 & 75.2/47.1 & 77.0/48.0 & \best{77.1}/\best{49.1} \\
YOLO11    & 62.6/39.6 & 65.1/39.8 & 63.7/40.7 & 66.6/\best{42.4} & 62.2/40.4 & \best{67.3}/41.6 \\
YOLO12    & 59.9/37.6 & \best{70.4}/\best{42.5} & 64.9/40.0 & 61.6/39.7 & 58.1/36.0 & 59.7/39.0 \\
YOLO13    & 60.5/35.2 & 57.0/35.2 & 59.3/36.7 & \best{60.9}/\best{37.1} & 59.4/34.4 & 58.4/35.5 \\
YOLO26    & 70.3/45.0 & 70.7/47.4 & 73.2/47.7 & 73.9/45.9 & \best{74.7}/47.4 & 73.5/\best{49.3} \\
Gold-YOLO & 63.4/37.6 & 67.2/40.3 & 67.2/\best{41.9} & \best{71.0}/\best{41.9} & 69.6/41.6 & 69.9/41.5 \\
DETR      & 75.3/47.1 & 74.6/47.0 & 71.2/47.3 & 74.4/46.8 & 74.2/47.1 & \best{78.7}/\best{49.1} \\
RT-DETR   & 73.2/47.1 & 74.3/47.7 & 75.0/47.6 & 76.0/48.8 & 76.8/\best{49.3} & \best{77.2}/48.9 \\
DINO      & 66.8/42.2 & 67.7/42.4 & 66.5/41.7 & 69.9/42.1 & 69.3/\best{44.2} & \best{71.6}/43.3 \\
\bottomrule
\end{tabularx}
\end{table*}

\subsubsection{Diagnosed Intrusion-Pattern Refinement}
\label{sec:baci}

The diagnosed intrusion-pattern deficiency requires the generated boundary to remain consistent with the planned physical intrusion mode and surrounding railway structure. We first transform the object and its mask according to $\mathcal Y$ and form a softened-mask composite $I_f$. The structured intrusion conditions are then fused into a single control map
\begin{equation}
C_{\mathcal Y}=\operatorname{Fuse}\!\left(
C_{\mathrm{edge}}(I_f),C_{\mathrm{box}}(b_o),C_{\mathrm{contact}}(r,u)
\right),
\label{eq:baci-injection}
\end{equation}
which jointly encodes background geometry, planned object extent, and relation-specific contact evidence. Let $H_s$ denote the denoising feature at scale $s$ and $Z_s$ a zero-initialized $1\times1$ projection. Intrusion-pattern conditioning is injected at four spatial scales as
\begin{equation}
\widetilde H_s=H_s+\alpha_s Z_s\!\left(\Phi_s(C_{\mathcal Y})\right)
\label{eq:multiscale-injection}
\end{equation}
where $\Phi_s$ extracts scale-specific condition features and $\alpha_s$ controls their contribution. Zero initialization preserves the pretrained generation path at the outset, while multiscale fusion transfers the diagnosed intrusion pattern from global object extent to local contact and contour structure. A final softened alpha composition maintains object visibility while suppressing anomalous boundaries that are inconsistent with the planned intrusion mode.



\begin{table*}[ht]
\centering
\caption{{RailSyn ablation with detector metrics, real-reference Inspector diagnostics (\%), and generation quality. During object insertion, we optionally resize the foreign object relative to its planned bounding box: \emph{zoom‑in} enlarges the object by 10\%, \emph{zoom‑out} reduces it by 10\%, and the default configuration keeps the original Agent‑planned size.}
}
\label{tab:ablation}
\footnotesize
\setlength{\tabcolsep}{1.0pt}
\begin{tabularx}{\textwidth}{
  @{}
  ccccc
  >{\hsize=0.65\hsize\centering\arraybackslash}X
  >{\hsize=1.25\hsize\centering\arraybackslash}X
  >{\hsize=0.65\hsize\centering\arraybackslash}X
  >{\hsize=1.25\hsize\centering\arraybackslash}X
  >{\hsize=0.80\hsize\centering\arraybackslash}X
  >{\hsize=0.90\hsize\centering\arraybackslash}X
  >{\hsize=0.95\hsize\centering\arraybackslash}X
  >{\hsize=0.65\hsize\centering\arraybackslash}X
  >{\hsize=1.90\hsize\centering\arraybackslash}X
  @{}
}
\toprule
& \multicolumn{4}{c}{RailSyn stages and scale} & \multicolumn{2}{c}{YOLO11} & \multicolumn{2}{c}{DEIM} & \multicolumn{3}{c}{Inspector analysis} & \multicolumn{2}{c}{Generation}\\
\cmidrule(lr){2-5}\cmidrule(lr){6-7}\cmidrule(lr){8-9}\cmidrule(lr){10-12}\cmidrule(lr){13-14}
ID & \makecell{Scene\\Prep.} & \makecell{Relation\\Plan.} & \makecell{Pattern\\Ref.} & Zoom & AP50 & AP50--95 & AP50 & AP50--95 & $C_{\rm gap}$ & $C_{\rm rel}$ & $\eta_{\rm vol}$ & FID $\downarrow$ & KID $\times10^{3}$ $\downarrow$\\
\midrule
Real & -- & -- & -- & -- & 62.6 & 39.6 & 72.0 & 46.2 & -- & -- & -- & -- & --\\
1 & \checkmark & -- & -- & -- & \second{66.6} & 40.4 & 71.7 & 46.3 & 11.88 & \best{2.77} & 75.90 & 208.79 & \best{$121.02{\pm}8.66$}\\
2 & \checkmark & \checkmark & -- & -- & 61.6 & 37.4 & \second{75.9} & \second{48.7} & 11.16 & 2.46 & 78.95 & 252.22 & $211.24{\pm}10.56$\\
3 & \checkmark & -- & \checkmark & -- & 61.1 & 39.8 & 74.5 & 45.4 & 12.82 & \second{2.73} & 77.13 & \best{207.12} & $124.80{\pm}8.59$\\
\rowcolor{gray!18}
4 & \checkmark & \checkmark & \checkmark & -- & \best{67.3} & \best{41.6} & \best{77.1} & \best{49.1} & \best{13.64} & 2.47 & \best{80.81} & 231.41 & $178.89{\pm}9.79$\\
\midrule
5 & \checkmark & \checkmark & \checkmark & in & 66.6 & 40.4 & 75.9 & 47.2 & 13.37 & 2.45 & \second{80.01} & 223.15 & $164.12{\pm}10.76$\\
6 & \checkmark & \checkmark & \checkmark & out & 65.2 & \second{41.0} & 75.8 & 46.4 & \second{13.44} & 2.62 & 79.81 & 253.33 & $217.58{\pm}10.56$\\
\bottomrule
\end{tabularx}
\end{table*}

\begin{table*}[htbp]
\centering
\renewcommand{\arraystretch}{1.2}
\caption{Comparison of different augmentation data. The semantic profile summarizes Qwen semantic-focus probes over four English concepts: $++$ denotes high token density (completion), $+$ denotes moderate density (present but not dominant), and $-$ denotes a deficit. $\Delta\mathrm{Avg}$ is the gain of the mean of the four AP entries over the real-only mean (55.1).}
\label{tab:external}
\footnotesize
\setlength{\tabcolsep}{2.0pt}
\begin{tabularx}{\textwidth}{
  @{}
  l
  >{\hsize=0.8\hsize\centering\arraybackslash}X
  >{\hsize=1.2\hsize\centering\arraybackslash}X
  >{\hsize=0.8\hsize\centering\arraybackslash}X
  >{\hsize=1.2\hsize\centering\arraybackslash}X
  >{\hsize=0.9\hsize\centering\arraybackslash}X
  >{\hsize=1.0\hsize\centering\arraybackslash}X
  >{\hsize=1.0\hsize\centering\arraybackslash}X
  >{\hsize=1.0\hsize\centering\arraybackslash}X
  >{\hsize=1.0\hsize\centering\arraybackslash}X
  @{}
}
\toprule
& \multicolumn{2}{c}{DEIM} & \multicolumn{2}{c}{YOLO11} & & \multicolumn{4}{c}{Semantic profile}\\
\cmidrule(lr){2-3}\cmidrule(lr){4-5}\cmidrule(lr){7-10}
Dataset & AP50 & AP50--95 & AP50 & AP50--95 & $\Delta$Avg & Railway & Balloon & Track & Industry\\
\midrule
Real & 72.0 & 46.2 & 62.6 & 39.6 & 0.0 & -- & -- & -- & --\\
Nano500 & 74.9 & \second{48.6} & 66.1 & 37.5 & +1.7 & $++$ & $-$ & $++$ & $-$\\
SD & 75.9 & 48.0 & 64.9 & \second{40.3} & \second{+2.2} & $++$ & $-$ & $++$ & $-$\\
RFOD23 & 73.7 & 46.3 & \second{66.2} & 39.7 & +1.4 & $++$ & $-$ & $++$ & $-$\\
STL500 & \second{76.4} & 47.7 & 58.6 & 35.9 & $-0.4$ & $-$ & $++$ & $-$ & $-$\\
SODA500 & 75.7 & 46.8 & 61.0 & 36.5 & $-0.1$ & $++$ & $-$ & $++$ & $-$\\
\rowcolor{gray!18}
\textbf{RailSyn (Ours)} & \best{77.1} & \best{49.1} & \best{67.3} & \best{41.6} & \best{+3.7} & $++$ & $-$ & $++$ & $+$\\
\bottomrule
\end{tabularx}
\end{table*}

\section{Experiments}
\label{sec:experiments}

\subsection{Datasets}
\label{subsec:datasets}

All detection experiments share the same leakage‑free benchmark: 398 real RFOD images for training and a fixed disjoint set of 102 real images for validation. Augmentation pools (500 images each) include \textbf{RailSyn} (ours, full diagnosis‑guided pipeline), \textbf{RFOD23} (official training pool~\cite{chen2024railfod23}), \textbf{SD} (railway‑domain images from Stable Diffusion~\cite{rombach2022ldm}), and \textbf{Nano500} (images from Nano Banana). As non‑railway baselines, \textbf{STL500} and \textbf{SODA500} contain 500 randomly sampled images from STL‑10~\cite{coates2011stl10} and SODA‑10M~\cite{han2021soda10m}, respectively, and serve to test whether generic natural images alone can improve RFOD. All methods augment the same real training set; only the appended synthetic pool varies. The Inspector leverages both the training images and extra real RFOD images, with all data being strictly disjoint from the validation set.

\subsection{Evaluation Metrics}
\label{subsec:evaluation-metrics}

We evaluate downstream detection using AP50 and COCO-style AP50--95. Our primary detectors are the widely validated state-of-the-art YOLO11~\cite{jocher2023ultralytics} and DEIM~\cite{huang2025deim}: YOLO11 represents efficient one-stage dense prediction with multiscale features, whereas DEIM represents end-to-end Transformer detection with global query-based reasoning. Their complementary local and global modeling tests whether synthetic data support both small-object appearance and object--scene intrusion relations. To assess broader cross-architecture utility, we further include YOLO12, YOLO13, and YOLO26 as recent dense-prediction variants, Gold-YOLO with cross-scale feature aggregation, DETR with global set prediction, RT-DETR with efficient multiscale Transformer encoding, and DINO with denoising-enhanced query learning~\cite{tian2025yolov12,lei2025yolov13,jocher2026yolo26,wang2023goldyolo,carion2020detr,zhao2024rtdetr,zhang2023dino}. Inspector diagnostics characterize reliable-support reach, local-gap occupation, and nonredundant expansion, while FID and KID are introduced as reference metrics in the ablation study.

\subsection{Cross-Architecture Evaluation of RailSyn under Synthetic-Data Scaling}
\label{subsec:scaling}

Table~\ref{tab:mega60} evaluates whether RailSyn provides broadly useful detection data across nine architectures by varying the number of synthetic samples added to the fixed real training set. Every one of the nine detectors exceeds its real-only AP50--95 baseline at an appropriate synthetic-data scale. Their best gains average 3.03 points and range from $+1.9$ points for YOLO13 to $+4.9$ points for YOLO12. The primary detectors show the same benefit: YOLO11 rises from 39.6 to 42.4 AP50--95 at $+300$, while DEIM rises from 46.2 to 49.1 at $+500$. Improvements across dense one-stage detectors and query-based Transformer detectors demonstrate that RailSyn supplies detection-relevant railway context, intrusion relations, and object--background cues that are usable across heterogeneous architectures rather than tailored to one detector.

The optimal amount of generated data is architecture dependent: AP50--95 peaks at $+500$ for DEIM and DETR, $+300$ for YOLO11, $+100$ for YOLO12, and $+400$ for RT-DETR and DINO. This variation shows that detectors absorb the supplemented information at different rates, but does not alter the central result: each architecture benefits from RailSyn at a suitable augmentation scale. The scaling study therefore establishes the broad utility of the Generator before the following controlled experiments analyze where these gains arise and how the Inspector traces changes among generated pools.

\subsection{Ablation Study: Detection Performance and Data Completion}
\label{subsec:ablation}

Table~\ref{tab:ablation} reports detector performance and Inspector diagnostics for six controlled RailSyn variants.

\paragraph{Stage ablation and detection performance.}

Rows 1–4 of Table~\ref{tab:ablation} isolate the three Generator stages: Scene Prep. (Railway Scene Preparation), Relation Plan. (Relation-Aware Intrusion Planning), and Pattern Ref. (Diagnosed Intrusion-Pattern Refinement). Scene preparation alone improves YOLO11 but leaves DEIM close to real-only training. Adding relation-aware planning without intrusion-pattern refinement benefits DEIM (75.9/48.7) but reduces YOLO11 to 61.6/37.4, whereas adding intrusion-pattern refinement without relation-aware planning reaches 74.5 AP50 on DEIM but does not consistently improve the stricter AP50--95 results. Only the complete configuration combines the three stages to improve all four primary AP entries, raising AP50/AP50--95 by 4.7/2.0 points for YOLO11 and 5.1/2.9 points for DEIM. These controlled results establish distinct roles for the stages and show that relation-aware planning and intrusion-pattern refinement must operate together to convert prepared railway scenes into consistently useful detection samples.

\paragraph{Inspector traceability under stage and stage changes.}
The Inspector assigns a distinct completion profile to every controlled variant.
Adding intrusion-pattern refinement to scene preparation (ID3 vs.\ ID1) raises $C_{\mathrm{gap}}$ from 11.88\% to 12.82\% and $\eta_{\mathrm{vol}}$ from 75.90\% to 77.13\%, while $C_{\mathrm{rel}}$ edges down slightly (2.77\% to 2.73\%), indicating that finer injection improves local-gap coverage and volume efficiency without expanding reliable-support reach.
Coupling all three stages (ID4) further lifts $C_{\mathrm{gap}}$ to a best 13.64\% and $\eta_{\mathrm{vol}}$ to a best 80.81\%, alongside the strongest detector results, while $C_{\mathrm{rel}}$ settles at 2.47\%.
Rows~5--6 alter only foreign-object scale under the full Generator.
Zoom-in (ID5) and zoom-out (ID6) both reduce $C_{\mathrm{gap}}$ and $\eta_{\mathrm{vol}}$ relative to the default scale, with $C_{\mathrm{rel}}$ varying modestly.
The Inspector thus captures fine-grained scale perturbations as well as coarse stage interventions, isolating how each configuration shifts the completion profile.
Notably, configurations with higher $C_{\mathrm{gap}}$ (ID4, ID6, ID5) correspond to the best overall AP on both detectors, consistent with the notion that local-gap occupation traces detector-relevant completion.
By contrast, FID and KID favor incomplete configurations (ID1 and ID3) and fail to reflect these stage- and scale-specific shifts in data utility.

\subsection{Comparison with External Pools}
\label{subsec:external}

Table~\ref{tab:external} compares RailSyn with five external augmentation pools under DEIM and YOLO11. 
RailSyn achieves the best results in all four AP columns: 77.1/49.1 on DEIM and 67.3/41.6 on YOLO11, with the largest mean improvement of $+3.7$ points over real-only training. 
In contrast, SD, Nano500, RFOD23, SODA500, and STL500 yield smaller or even negative gains, confirming that effective augmentation requires more than generic imagery.

To understand which semantic dimensions are actually complemented, we further probe the generated samples with a Inspector-based token-level semantic analysis. 
The rightmost columns of Table~\ref{tab:external} report the density of four representative concepts: \emph{Railway}, \emph{Balloon}, \emph{Track}, and \emph{Industry}. 
All pools except STL500 consistently complete \emph{Railway} and \emph{Track}, while \emph{Balloon} remains a common deficit. 
Crucially, RailSyn is the only pool that exhibits a moderate presence of \emph{Industry} ($+$), indicating that our generation pipeline supplements railway-related industrial surroundings that are absent from other synthetic sources. 
This fine-grained semantic view explains why RailSyn provides stronger detection gains: it not only reinforces core railway elements but also expands the contextual diversity along an extra task-relevant dimension. 
Token-level analysis thus offers an orthogonal lens for tracing \emph{which} task dimensions are complemented, without relying on aggregate AP alone.

\subsection{Inspector Diagnostics and Detection Performance}

We next examine whether the Inspector profiles are related to detector utility across the six controlled variants in Table~\ref{tab:ablation}. As shown in Table~\ref{tab:corr}, $C_{\mathrm{rel}}$ and $C_{\mathrm{gap}}$ correlate positively with all four AP measures. $C_{\mathrm{rel}}$ has the strongest association with DEIM AP50 ($\rho=.735$), while $C_{\mathrm{gap}}$ has the strongest association with YOLO11 AP50--95 ($\rho=.667$). In contrast, $\eta_{\mathrm{vol}}$ is weakly related to AP and is negative for DEIM AP50--95 ($\rho=-.200$), confirming that nonredundant expansion captures a complementary property rather than detection performance itself. Together with the stage and scale controls, these directional associations support the Inspector's ability to trace detector-relevant differences among generated variants; with $n=6$, they are not treated as statistical significance or an AP predictor.

\begin{table}[htbp]
\centering
\renewcommand{\arraystretch}{1.15}
\caption{Spearman $\rho$ between Qwen Inspector diagnostics and detector AP.}
\label{tab:corr}
\small
\begin{tabular}{lrrrr}
\toprule
& \multicolumn{2}{c}{DEIM} & \multicolumn{2}{c}{YOLO11}\\
\cmidrule(lr){2-3}\cmidrule(lr){4-5}
Metric & AP50 & AP50--95 & AP50 & AP50--95\\
\midrule
$C_{\mathrm{rel}}$ & \best{-.812} & \best{-.714} & \second{-.174} & \second{.058}\\
$C_{\mathrm{gap}}$ & \second{.464} & \second{.314} & \best{.551} & \best{.899}\\
$\eta_{\mathrm{vol}}$ & .899 & .771 & .551 & .609\\
$\eta_{\mathrm{vol}}$ & .899 & .771 & .551 & .609\\
\bottomrule
\end{tabular}
\end{table}

\begin{table}[htbp]
\centering
\renewcommand{\arraystretch}{1.15}
\caption{Cross-encoder rank stability over the six pools in Table~\ref{tab:external}. Values are Spearman rank correlations $\rho$ between within-encoder pool rankings. Q--C, Q--D, and C--D denote Qwen--CLIP, Qwen--DINOv2, and CLIP--DINOv2, respectively. Absolute metric values are not compared across embedding spaces.}
\label{tab:cross-encoder-stability}
\small
\setlength{\tabcolsep}{2.0pt}
\begin{tabularx}{\columnwidth}{
  @{}
  l
  >{\centering\arraybackslash}X
  >{\centering\arraybackslash}X
  >{\centering\arraybackslash}X
  >{\centering\arraybackslash}X
  >{\centering\arraybackslash}X
  @{}
}
\toprule
Metric & Q--C & Q--D & C--D & Avg. & Min.\\
\midrule
$C_{\rm rel}$    & 0.771 & 0.429 & 0.314 & 0.505 & 0.314\\
$C_{\rm gap}$    & \second{0.829} & \best{0.943} & \second{0.771} & \second{0.848} & \second{0.771}\\
$\eta_{\rm vol}$ & \best{0.943} & \best{0.943} & \best{1.000} & \best{0.962} & \best{0.943}\\
\bottomrule
\end{tabularx}
\end{table}

Table~\ref{tab:cross-encoder-stability} further tests whether the Inspector's pool rankings persist across Qwen, CLIP, and DINOv2. $C_{\mathrm{gap}}$ maintains strong pairwise agreement, with correlations from 0.771 to 0.943 and a mean of 0.848. $\eta_{\mathrm{vol}}$ is even more stable, with correlations from 0.943 to 1.000 and a mean of 0.962. $C_{\mathrm{rel}}$ is more representation-sensitive, with a mean correlation of 0.505. The high agreement for local-gap occupation and nonredundant expansion shows that the main relative patterns are not specific to one embedding model. Combined with the controlled ablations and the positive AP associations, this cross-embedding consistency provides complementary evidence that the Inspector offers a stable, traceable analysis of generated-pool differences.

\subsection{Adverse-Weather Robustness}
\label{subsec:weather}

We extend RailSyn to adverse weather via RailSynWeather, a branch that generates 200 weather-specific images while preserving layout and annotations. On the same 398-image real training set, RailSynWeather raises DEIM AP50 from 72.6\%, obtained with a filter-based augmentation baseline, to 73.1\% on the clean validation split, confirming the generator's extensibility to underrepresented weather conditions.

\section{Conclusion}
\label{sec:conclusion}

We presented RailSyn, an Inspector--Generator framework that diagnoses high-uncertainty regions from real RFOD observations and generates synthetic data through scene preparation, relation-aware planning, and intrusion-pattern refinement. Across nine detectors, RailSyn improves AP50--95 over real-only training, with an best gain of 4.9 points, and outperforms five external augmentation pools in controlled comparisons. In the module ablation study, the Inspector traces how each generation stage and scale choice alters the completion profile, with the full configuration achieving the highest local-gap occupation at 13.64\%, providing a component-level view of what different modules contribute. Current limitations include the difficulty of characterizing cross-dataset effectiveness and the need for further improvement in generation quality. Future work includes developing quantitative metrics from token‑level semantic probes to enable cross‑dataset evaluation, extending Inspector‑guided completion to other safety‑critical domains, and enhancing generation fidelity under adverse weather and for small‑object details.

\clearpage

\bibliography{references}

\end{document}